\documentclass{article}
\PassOptionsToPackage{numbers, sort&compress}{natbib}
\usepackage[preprint]{neurips_2021}
\usepackage[colorlinks=true,linkcolor=red,anchorcolor=red,citecolor=green,urlcolor=blue]{hyperref}
\usepackage{url}
\usepackage{graphicx}
\usepackage{subfigure} 
\usepackage{amsthm,amsmath,amssymb}
\usepackage{booktabs}
\usepackage{mathrsfs}
\usepackage{bm}
\usepackage{multirow}
\usepackage{cleveref}
\usepackage{color}
\usepackage{xcolor}
\usepackage{url}
\usepackage{wrapfig}
\usepackage{graphicx}
\usepackage{amsmath}
\usepackage{amssymb}
\makeatletter
\newcommand{\thickhline}{%
    \noalign {\ifnum 0=`}\fi \hrule height 1pt
    \futurelet \reserved@a \@xhline
}

\title{CAT: Cross Attention in Vision Transformer}

\begin{document}

\author{
  Hezheng~Lin$^{12}$\thanks{Interns at MMU, KuaiShou Inc.} 
  \quad Xing~Cheng$^1$\footnotemark[1] 
  \quad Xiangyu~Wu$^1$\thanks{Corresponding author.} 
  \quad Fan~Yang$^1$ \\
  \textbf{ Dong~Shen$^1$
    \quad Zhongyuan~Wang$^1$
    \quad Qing~Song$^2$ 
    \quad Wei~Yuan$^1$}  \\
  [0.2cm]
  $^1$MMU, KuaiShou Inc. \quad$^2$Beijing University of Posts and Telecommunications, China\\
  [0.1cm]
  {$^1$ \tt\small \{linhezheng,chengxing03,wuxiangyu,yangfan,shendong,}\\
  {\tt\small wangzhongyuan,yuanwei05\}@kuaishou.com}\\
  {$^2$ \tt\small priv@bupt.edu.cn} 

}

\maketitle

\begin{abstract}
Since Transformer has found widespread use in NLP, the potential of Transformer in CV has been realized and has inspired many new approaches. However, the computation required for replacing word tokens with image patches for Transformer after the tokenization of the image is vast(e.g., ViT), which bottlenecks model training and inference. In this paper, we propose a new attention mechanism in Transformer termed Cross Attention, which alternates attention inner the image patch instead of the whole image to capture local information and apply attention between image patches which are divided from single-channel feature maps to capture global information. Both operations have less computation than standard self-attention in Transformer. By alternately applying attention inner patch and between patches, we implement cross attention to maintain the performance with lower computational cost and build a hierarchical network called Cross Attention Transformer(CAT) for other vision tasks. Our base model achieves state-of-the-arts on ImageNet-1K, and improves the performance of other methods on COCO and ADE20K, illustrating that our network has the potential to serve as general backbones. The code and models are available at \url{https://github.com/linhezheng19/CAT}.
\end{abstract}

\section{Introduction}
\label{introduction}
With the development of deep learning and the application of convolutional neural networks\cite{lecun1995convolutional}, computer vision tasks have improved tremendously. Since 2012, CNN has dominated CV for a long time, as a crucial feature extractor in various vision tasks, and as a task branch encoder in other tasks. A variety of CNN-based networks\cite{krizhevsky2012imagenet,simonyan2014very,szegedy2015going,ioffe2015batch,howard2017mobilenets,zhang2018shufflenet,he2016deep,xie2017aggregated,gao2019res2net,tan2019efficientnet} have different improvements and applications, and various downstream tasks also have these multiple methods, such as object detection\cite{ren2015faster,lin2017focal,lin2017feature,tian2019fcos,yolov5,bochkovskiy2020yolov4,redmon2018yolov3,liu2016ssd,ghiasi2019fpn}, semantic segmentation\cite{zhao2017pyramid,sun2019deep,yuan2019object,yu2015multi,chen2014semantic,chen2017deeplab,chen2017rethinking,chen2018encoder}.

Lately, Transformer\cite{vaswani2017attention}, as a new network structure, has achieved significant results in NLP. Benefiting from its remarkable ability to extract global information, it also solves the problem that sequence models hard to be parallelized such as RNN\cite{zaremba2014recurrent} and LSTM\cite{hochreiter1997long}, making the development of the NLP an essential leap, and also inspiring computer vision tasks. 

\begin{figure}[h]
\centering
\includegraphics[scale=0.3]{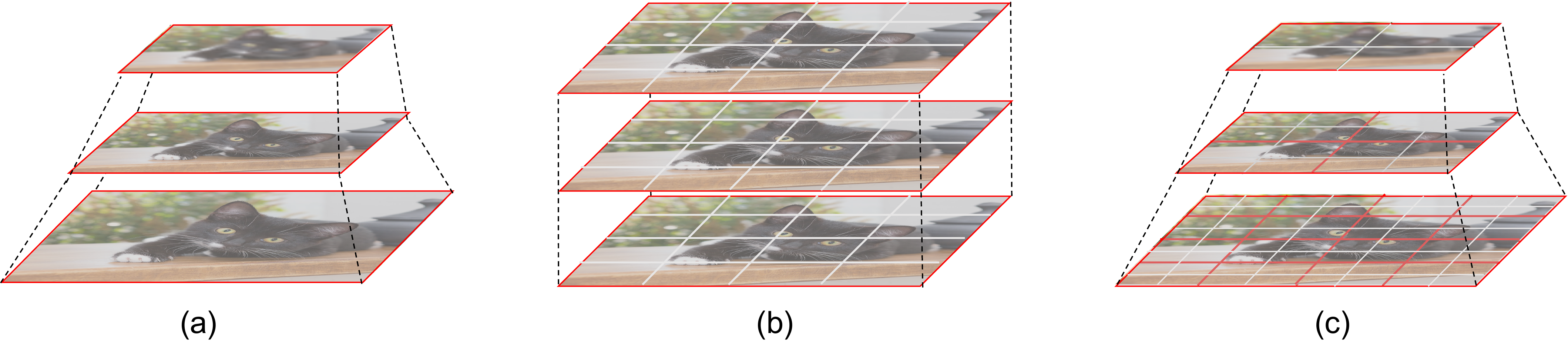}
\caption{Hierarchical network. (a)~Hierarchical networks based on CNN, different stage generates feature with variety scale. (b)~Hierarchical network based on Transformer(e.g., Vit), all features are same in shape. (c)~Hierarchical networks of CAT(ours), with characteristics of CNN hierarchy network.}
\label{hierarchical}
\end{figure}

Recent works\cite{wang2021pyramid,dosovitskiy2020image,chen2021crossvit,wu2021cvt,liu2021swin,han2021transformer,touvron2020training,zhou2021deepvit,touvron2021going,yuan2021tokens} introduces Transformer into the computer vision as an image extractor. However, the length of the text sequence is fixed in NLP which leads to a decrease in the ability of the Transformer to process images, since the resolution of inputs are variational in different task. In processing images with Transformer, one naive approach is to treat each pixel as a token for global attention similar to work tokens. The iGPT\cite{chen2020generative} demonstrates that the computation brought by this is tremendous. Some works(e.g., ViT, iGPT) take a set of pixels in a region as a token, which reduces the computation to a certain extent. However, the computational complexity increases dramatically as the input size increases(Formula~\ref{eqmsa}), and the feature maps generated in these methods are of the same shape(Figure~\ref{hierarchical}\textcolor{red}{(b)}), making these methods unsuitable for use as the backbone of subsequent tasks.

In this paper, we are inspired by the local feature extraction capabilities of CNN, we adopt attention between pixels in one patch to simulate the characteristics of CNN, reducing the computation that increases exponentially with the input size to that is exponentially related to the patch size. Meanwhile, as Figure~\ref{cpsa} shown, to consider the overall information extraction and communication of the picture, we devised a method of performing attention on single-channel feature maps. Compared with the attention on all channels, there is a significant reduction in the computation as Formula~\ref{eqmsa}, and \ref{eqncpsa} demonstrated. Cross attention is performed by alternating the internal attention of the patch and the attention of single-channel feature maps. We can build a powerful backbone with the Cross Attention to generate feature maps of different scales, which satisfies the requirements of different granular features of downstream tasks, as Figure~\ref{hierarchical} shown. We introduce global attention without increasing computation or a small increase in computation, which is a more reasonable method to joint features of Transformer and CNN.

Our base model achieves 82.8\% of top-1 accuracy on ImageNet-1K, which is comparable with the current CNN-based network and Transformer-based network of state-of-the-arts. Meanwhile, in other vision tasks, our CAT as the backbone in object detection and semantic segmentation methods can improve their performance.

The features of Transformer and CNN complement each other and that it is our long-term goal to combine them more efficiently and perfectly to take advantage of both. Our proposed CAT is a step in that direction, and hopefully, there will be better developments in that direction.

\section{Related work}
\label{related-work}

\paragraph{CNN/ CNN-based network} CNN has the characteristics of shared weights, translation, rotation invariance, and locality, which has made great achievements in computer vision instead of the multi-layer perceptron and has become the standard network in vision tasks in last decade. As the first CNN network to achieve great success in computer vision, AlexNet laid the foundation for the later development of the CNN-based network, and \cite{simonyan2014very,he2016deep,xie2017aggregated,gao2019res2net,tan2019efficientnet,brock2021high} for performance improvement have become the choice as the backbone in vision tasks. The Inceptions\cite{szegedy2015going,ioffe2015batch,szegedy2016rethinking,szegedy2017inception,chollet2017xception}, MobileNets\cite{howard2017mobilenets,sandler2018mobilenetv2,howard2019searching}, and ShuffleNets\cite{zhang2018shufflenet,ma2018shufflenet} for efficiency improvement are also alternatives in tasks required speed of inference.

\paragraph{Global attention in Transformer-based network} Transformer is proposed in NLP for machine translation, where the core multi-head self-attention(MSA\cite{vaswani2017attention}) mechanism is vital in extracting the characteristics of relationships between words at multiple levels. As the first few Transformer-based backbones, ViT\cite{dosovitskiy2020image} and Deit\cite{touvron2020training} divide the image into patches (patch size is 16$\times$16). One patch flattened as a token, and CLS-Token\cite{devlin2018bert} is introduced for classification. Both CvT\cite{wu2021cvt} and CeiT\cite{yuan2021incorporating} introduce the convolutional layer to replace the linear projection of QKV\cite{vaswani2017attention}. CrossViT\cite{chen2021crossvit} integrates global features of different granularity through dividing images into different sizes of patches for two branches. However, these methods put all patches together for MSA, only focusing on the relationship between different patches, and as the input size increases, the computational complexity increases dramatically as demonstrated in Formula \ref{eqmsa}, which is difficult to be applied to the vision tasks requiring large resolution input.

\paragraph{Local attention in Transformer-based network} The relationship between internal information of the patch is vital in vision\cite{lowe1999object,brendel2019approximating}. Recently, TNT\cite{han2021transformer} divides each patch into smaller patches. Through the proposed TNT block, the global information and the information inner the patch are captured. Swin\cite{liu2021swin} treats each patch as a window to extract the internal relevancy of the patch and shifted window is used to catch more features. However, the two methods have their problems. Firstly, to combine the global information interaction and local information interaction, the increase of computation could not be underestimated in \cite{han2021transformer}. Second, the interaction between local information interaction and adjacent patches lacks global information interaction in \cite{liu2021swin}. We propose a cross-patch self-attention block to effectively maintain global information interaction while avoiding the enormous increase of computation with the increase of the resolution of inputs.

\paragraph{Hierarchy networks and downstream tasks}Transformer has been used successfully for vision tasks\cite{carion2020end,zhu2020deformable,zheng2020rethinking,wang2020max} and NLP tasks\cite{devlin2018bert,yang2019xlnet,liu2019roberta,lan2019albert,sun2019ernie,joshi2020spanbert,liu2020k}. However, due to the consistent shape of input and output in typical Transformer, it is difficult to achieve the hierarchical structure similar to CNN-based networks\cite{he2016deep,xie2017aggregated,liu2016ssd,howard2019searching} which is significant in downstream tasks. FPNs\cite{lin2017feature,liu2018path,ghiasi2019fpn,tan2020efficientdet} combined with ResNet\cite{he2016deep} have become the standard paradigm in object detection. In semantic segmentation\cite{zhao2017pyramid,sun2019deep,yuan2019object,chen2018encoder}, the payramidal features are used to improve the performance. Recent PVT\cite{wang2021pyramid,liu2021swin} and Swin\cite{liu2021swin} reduce the resolution of feature in different stage similar to ResNet\cite{he2016deep}, which is also the method we used.

\section{Method}
\label{method}
\subsection{Overall architecture}

\begin{figure}
\centering
\subfigure[]{\includegraphics[scale=0.27]{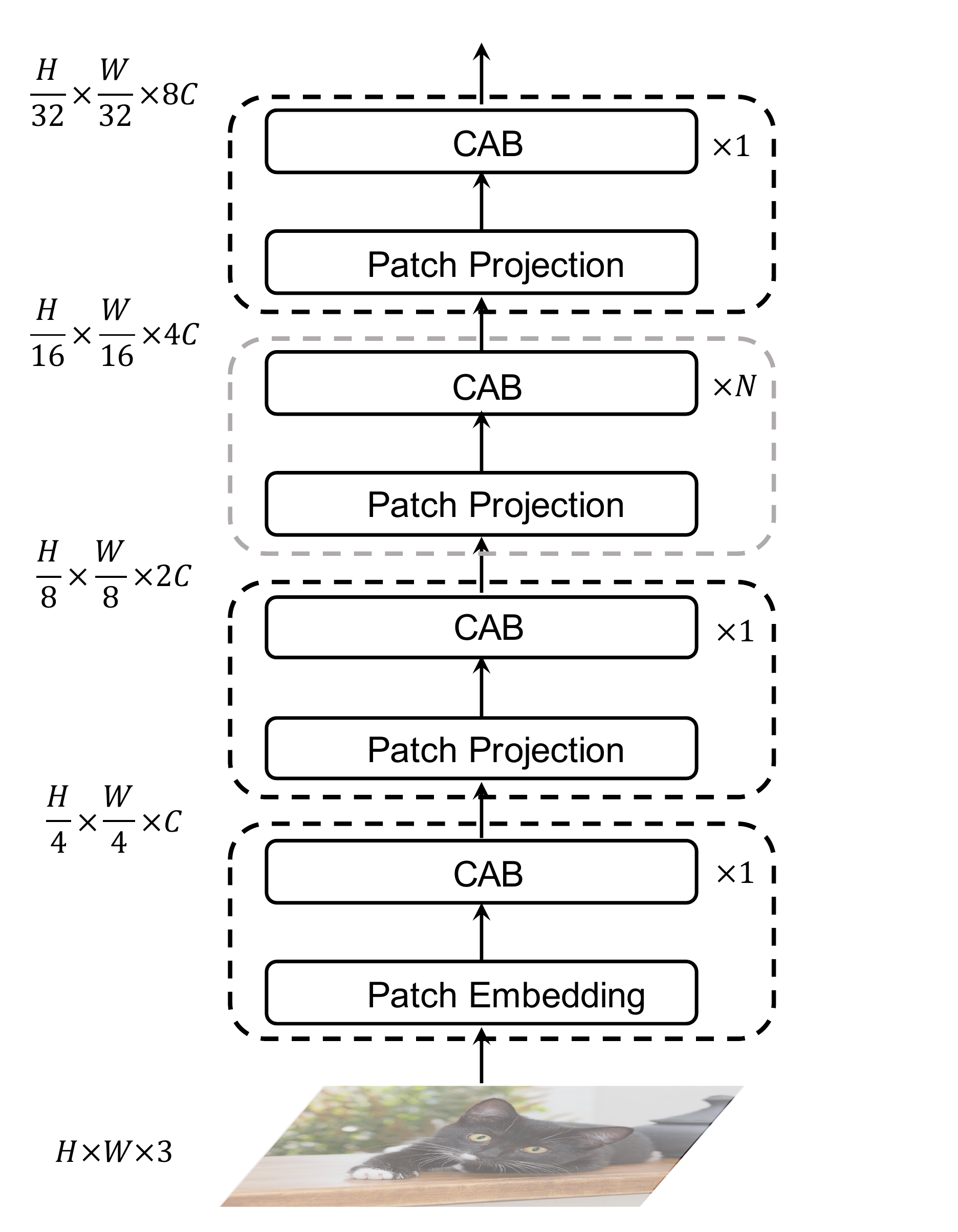}\label{architcheture}}
\subfigure[]{\includegraphics[scale=0.3]{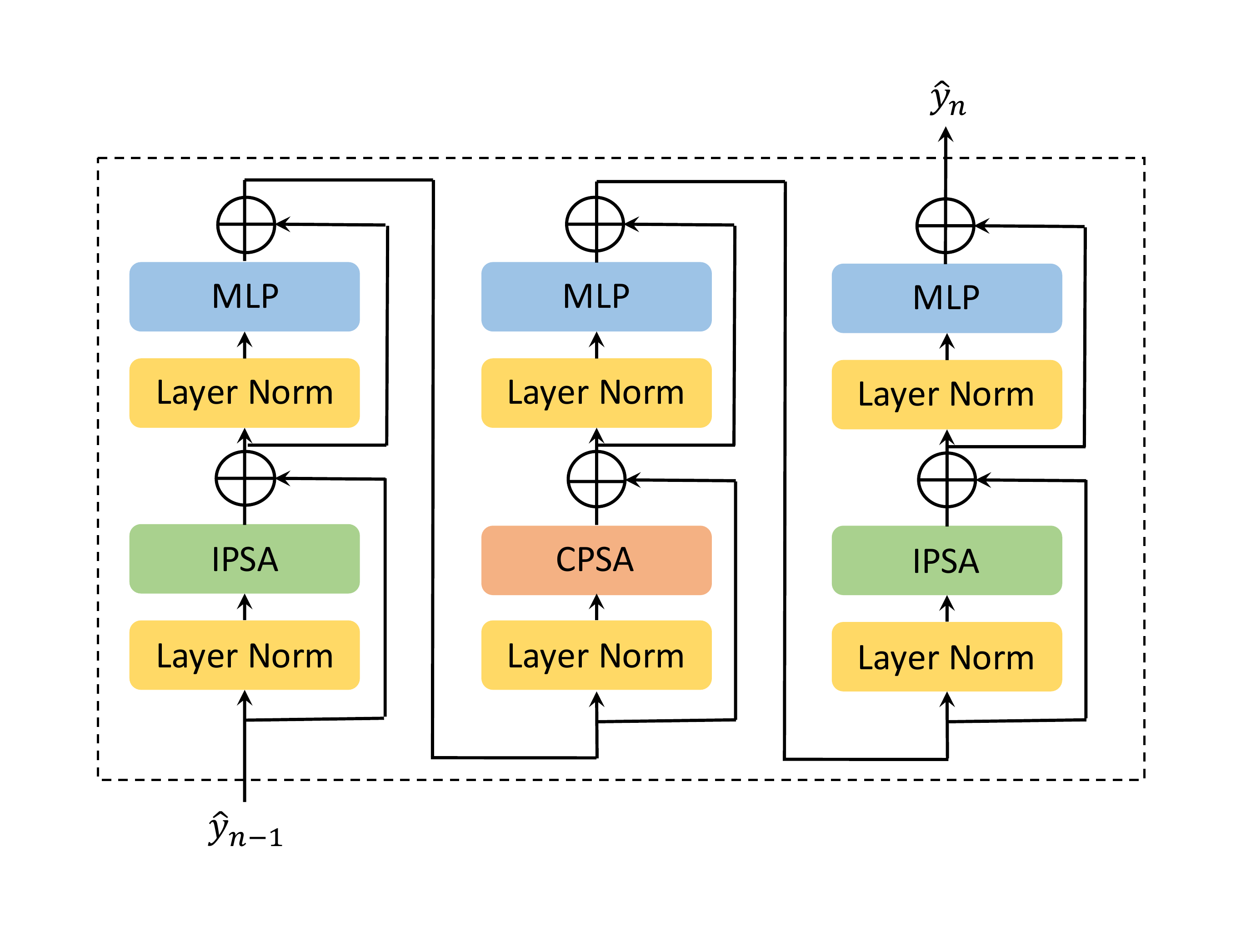}\label{cab}}
\caption{(a)~CAT architecture, at the third stage, the number of CABs varies with the size of model. (b)~Cross Attention Block(CAB), stacking IPSA and CPSA, both with LN\cite{ba2016layer}, MLP, and shorcut\cite{he2016deep}.}
\end{figure}

Our method aims to combine the attention within the patch and the attention between patches and build a hierarchical network by stacking basic blocks, which could be simplely applied in other vision tasks. As shown in Figure~\textcolor{red}{\ref{architcheture}}, firstly, we reduced the input image to $H_1=H$/$P, W_1=W/P$ (where $P=4$ in our experiments), and increase the number of channels to $C_1$ by referring to the patch processing mode in ViT\cite{dosovitskiy2020image} with patch embedding layer. Then, several CAT layers were used for feature extraction at different scales.

After the pretreatment above, the input image enters the first stage. At this point, the number of patches is $H_1/N \times W_1/N$, and the shape of the patch is $N\times N \times C_1$(where N is patch size after patch embedding layer). The shape of feature map output by stage1 is $H_1\times W_1 \times C_1$ donated as $F_1$. Then, enter the second stage, patch projection layer to execute space to depth operation, which performs pixel block with the shape of $2\times 2 \times C$ changes from the shape of $2\times 2\times C$ to the shape of $1\times 1\times 4C$, and then project to $1\times 1\times 2C$ through the linear projection layer. After entering several cross-attention blocks in the next stage and generating $F_2$ with a shape of $H_1/2 \times W_1/2 \times C_2$, the length and width of the feature map can be reduced by one time and the dimension is increased to double, similar to the operation in ResNet\cite{he2016deep}, which is also the practice in Swin\cite{liu2021swin}. After passing the four stages, we can get $\{F_1, F_2, F_3, F_4\}$, four feature maps of different scales and dimensions. Like typical CNN-based networks\cite{he2016deep,xie2017aggregated}, feature maps of different granularity can be provided for other downstream vision tasks.

\subsubsection{Inner-Patch Self-Attention Block}
In computer vision, each pixel needs a specific channel to represent its different semantic features. Similar to word tokens in NLP, the ideal is to take each pixel of feature map as a token (e.g., ViT, DeiT), but the computational cost is too enormous. As Formula~\ref{eqmsa} shows, the computational complexity increases exponentially with the resolution of the input image. For instance, in the conventional RCNN series\cite{girshick2014rich, girshick2015fast,ren2015faster,lin2017focal} of methods, the short edge of the input is at least 800 pixels, while the YOLO series\cite{bochkovskiy2020yolov4,wang2020scaled,yolov5} of papers also need images of more than 500 pixels. Most of the semantic segmentation methods\cite{zhao2017pyramid,sun2019deep,chen2014semantic} also need images with 512 pixels side lengths. The computation cost is at least 5 times higher than that of 224 pixels in pre-training phase.
\begin{eqnarray}
\label{eqmsa}
 &FLOPs_{MSA}=4HWC^2 + 2H^2W^2C
\end{eqnarray}

\begin{figure}[h]
\centering
\subfigure[]{\includegraphics[scale=0.35]{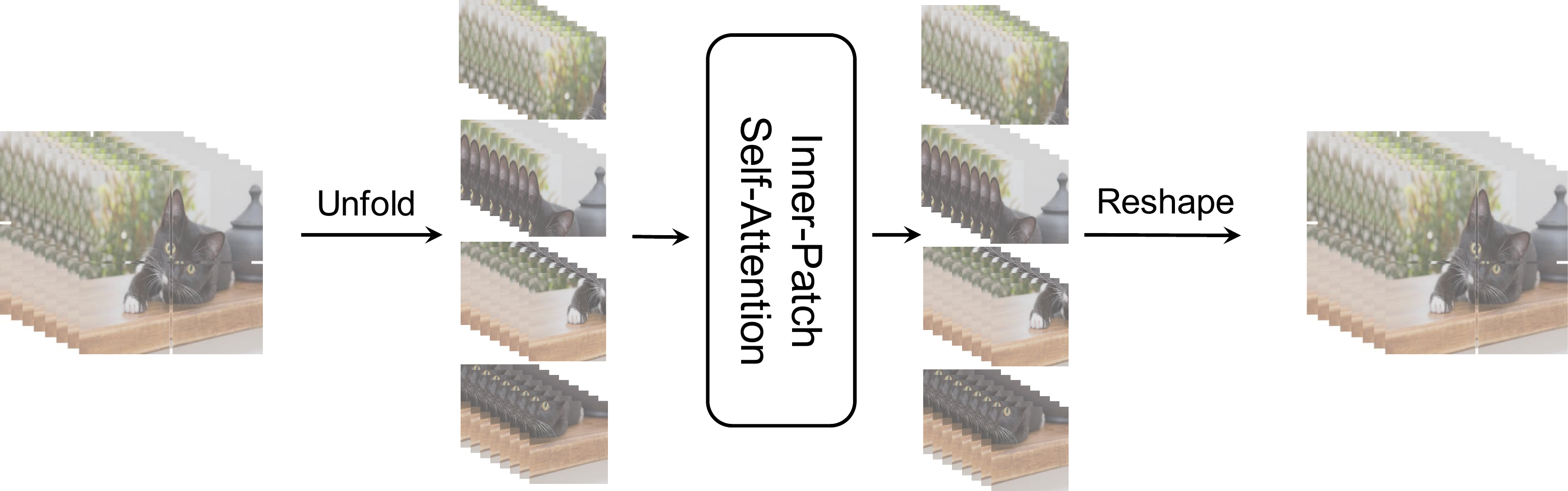}\label{ipsa}}
\subfigure[]{\includegraphics[scale=0.35]{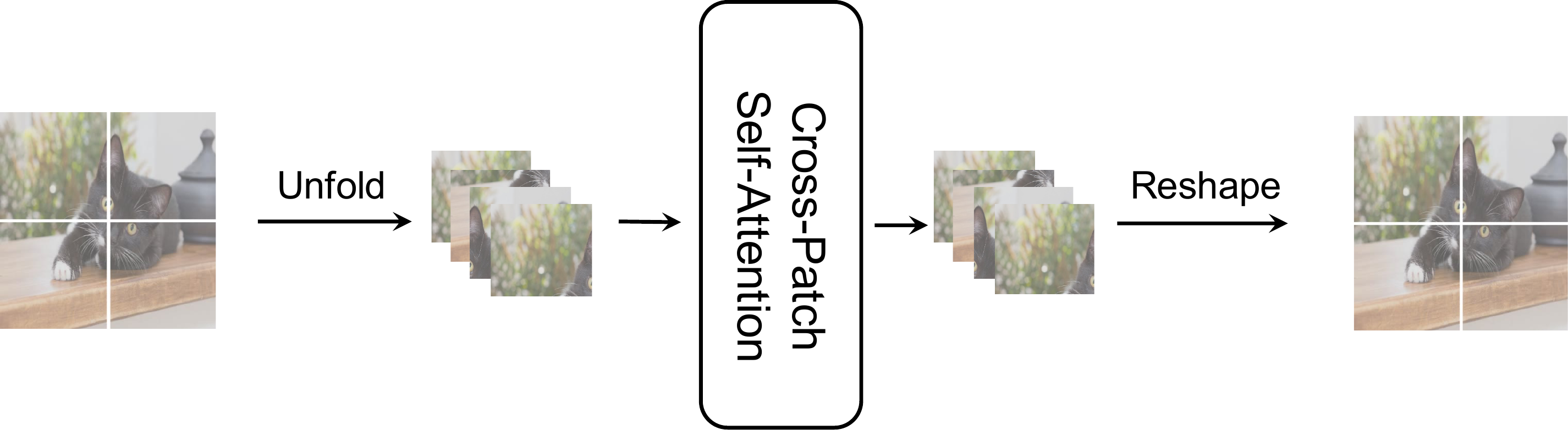}\label{cpsa}}
\caption{The pipeline of IPSA and CPSA. (a)~IPSA: unfold the all-channel inputs to $2\times$2, and stack them, after IPSA block, reshape to original shape. (b)~CPSA: unfold the single-channel input to $2\times2$ patches and stack them, after CPSA block, reshape to original shape.}
\end{figure}

Inspired by the characteristics of local feature extraction of CNN, we introduce the locality of convolution method in CNN into Transformer to conduct per-pixel self-attention in each patch called Inner-Patch Self-Attention(IPSA) as shown in Figure~\textcolor{red}{\ref{ipsa}}. We treat a patch as an attention scope, rather than the whole picture. At the same time, Transformer can generate different attention-maps according to inputs, which has a significant advantage over CNN with fixed parameters, which is similar to dynamic parameters in convolutional method, and it is be proved gainful in \cite{tian2020conditional}. \cite{han2021transformer} has revealed that attention between pixels is also vital. Our approach significantly reduces computation while taking into account the relationship between pixels in the patch. The formula of computation as follows:
\begin{eqnarray}
\label{eqnipsa}
 &FLOPs_{IPSA}=4HWC^2 + 2N^2HWC
\end{eqnarray}
where $N$ is patch size in IPSA. Compared with MSA in a standard Transformer, the computational complexity decreased from a quadratic correlation(Fornula~\ref{eqmsa}) with the $H\times W$ to a linear correlation with the $H\times W$. Assume that $H, W = 56$, $C = 96$, $N=7$, $FLOPs_{MSA} \approx 2.0~G$ following Formula~\ref{eqmsa}, and following Formula~\ref{eqnipsa}, $FLOPs_{IPSA} \approx 0.15~G$, which is much fewer.

\subsubsection{Cross-Patch Self-Attention Block}
Adding the attention mechanism between pixels only ensures that the interrelationships between pixels inner one patch be caught, but the information exchange of the whole picture is also quite crucial. In CNN-based networks, a stacked convolution kernel generally practiced expanding the receptive field. Dilated/Atrous Convolution\cite{yu2015multi} is proposed for larger receptive field, and the final receptive field expands to the whole picture is expected in practice. Transformer is naturally capable of capturing global information, but efforts like ViT\cite{dosovitskiy2020image} and Deit\cite{touvron2020training} are ultimately not the best resolution.

Each single-channel feature map naturally has global spatial information. We propose Cross-Patch Self-Attention, separating each channel feature map and dividing each channel into $H/N \times W/N$ patches and using self-attention to get global information in the whole feature map. This is similar to the depth-wise separable convolution used in Xception\cite{chollet2017xception} and MobileNet\cite{howard2017mobilenets}. The computation of our method could be computed as follows:
\begin{eqnarray}
\label{eqncpsa}
 &FLOPs_{CPSA} = 4N^2HWC+2(HW/N)^2C
\end{eqnarray}

where $N$ is patch size in CPSA, $H$, $W$ represent height and width of feature map respectively. The computational cost is fewer than ViT(Formula~\ref{eqmsa}) and other global attention based methods. Meanwhile, as shown in Figure~\textcolor{red}{\ref{cab}}, we combine with MobileNet\cite{howard2017mobilenets} design, stacking IPSA block and CPSA block to extract and integrate features between pixels in one patch and between patches in one feature map. Compared to the shifted window in Swin\cite{liu2021swin}, which is manually designed, difficult to implement, and has little ability to capture the global information, ours is reasonable and easier to comprehend. The $FLOPs_{CPSA}$ is about $0.1~G$ computed follow the Formula~\ref{eqncpsa} with the suppose same as the above section, which is much fewer than $2.0~G$ of MSA.

The multi-head self-attention mechanism is proposed in \cite{vaswani2017attention}. Each head can notice different semantic information between words in NLP. In computer vision, each head can notice different semantic information between image patches which is similar to channels in CNN-based networks. In the CPSA, we set the number of heads as patch size making the dimension of one head equal to patch size, which is useless to performance, as presented in Table~\ref{ablation}. So the single head is the default setting in our experiments.

\paragraph{Position encoding} We adopt relative position encoding in IPSA refer to \cite{hu2018relation,bao2020unilmv2,liu2021swin}, while for CPSA which conduct self-attention on the complete single-channel feature map, we add absolute position encoding to features which embedded in patch embedding layer, which could be formed as follows:

\begin{eqnarray}
\label{abpos}
 &y=Patch.Emb(x_{input})\\
 &y_{temp}=IPSA(y + ab.pos.)\\
 &y_{output}=IPSA(CPSA(y_{temp}))
\end{eqnarray}

where ab.pos. indicates that absolute position encoding, and Patch.Emb indicates that patch embedding layer in Table~\ref{model-configs}. Absolute position encoding is useful in CPSA to improve the performance, the results reported in Table~\ref{ablation2}.

\subsubsection{Cross Attention based Transformer}
Cross Attention block consists of two inner-patch self-attention blocks and a cross-patch self-attention block, as shown in Figure~\textcolor{red}{\ref{cab}}. CAT Layer is composed of several CABs, and each stage of the network is composed of a different number of layers and a patch embedding layer as shown in Figure~\textcolor{red}{\ref{architcheture}}, the pipeline of CAB is as follows:
\begin{eqnarray}
\label{eqncab}
 &\hat{y}_{temp1}=IPSA(LN(\hat{y}_{n-1})) + \hat{y}_{n-1} \\
 &\hat{y}_{temp2}=MLP(LN(\hat{y}_{temp1})) + \hat{y}_{temp1} \\
 &\hat{y}_{temp3}=CPSA(LN(\hat{y}_{temp2})) + \hat{y}_{temp2}\\
 &\hat{y}_{temp4}=MLP(LN(\hat{y}_{temp3})) + \hat{y})_{temp3} \\
 &\hat{y}_{temp5}=IPSA(LN(\hat{y}_{temp4})) + \hat{y}_{temp4} \\
 &\hat{y}_{n}=MLP(LN(\hat{y}_{temp5})) + \hat{y}_{temp5}
\end{eqnarray}
where $\hat{y}_{tempi}$ is an output of one block(e.g., IPSA, MLP) with LN. We compare convolution for patch embedding layer in \cite{dosovitskiy2020image}, where the convolution kernel size is set to P and the stride is also P, and slicing the inputs as \cite{yolov5}, the result is reported in Table~\ref{ablation}, both have the same performance. Our default setting is former. According to the number of CABs in stage3 and the dimension of patch projection layer, three models of different computational complexity are designed, which are CAT-T, CAT-S, and CAT-B with $1\times$, $2\times$, and $3\times$ of computation, respectively. Table~\ref{model-configs} details the configuration.

\begin{table*}[]
\small
\centering
\caption{Detailed configurations of CATs. down. rate indicates that down-sample rate at each stage. R indicates that dowm-sample rate at specific layer.}
\label{model-configs}
\addtolength{\tabcolsep}{-2pt}
\begin{tabular}{c|c|c|c|c} 
\thickhline
 & \begin{tabular}[c]{@{}c@{}}down. rate\end{tabular} & CAT-T  & CAT-S & CAT-B \\
\thickhline
\multirow{3}{*}{stage 1} & \multirow{3}{*}{\begin{tabular}[c]{@{}c@{}}4$\times$ \end{tabular}} & 
\begin{tabular}{c} Patch Embedding \\ R=4, Dim=64 \end{tabular} &
\begin{tabular}{c} Patch Embedding \\ R=4, Dim=96 \end{tabular} &
\begin{tabular}{c} Patch Embedding \\ R=4, Dim=96 \end{tabular}  \\
\cline{3-5}
& & $\begin{bmatrix}\text{IPSA, head 2,}\\\text{CPSA, head 1,}\\\text{IPSA, head 2,}\\\text{patch. sz. 7 $\times$7}\end{bmatrix}$ $\times$ 1   & 
$\begin{bmatrix}\text{IPSA, head 3,}\\\text{CPSA, head 1,}\\\text{IPSA, head 3,}\\\text{patch. sz. 7 $\times$7}\end{bmatrix}$ $\times$ 1    &
$\begin{bmatrix}\text{IPSA, head 3,}\\\text{CPSA, head 1,}\\\text{IPSA, head 3,}\\\text{patch. sz. 7 $\times$7}\end{bmatrix}$ $\times$ 1     \\
\hline
\multirow{3}{*}{stage 2}  & \multirow{3}{*}{\begin{tabular}[c]{@{}c@{}}8$\times$\\\end{tabular}} & 
\begin{tabular}{c} Patch Projection \\ R=2, Dim=128 \end{tabular} &
\begin{tabular}{c} Patch Projection \\ R=2, Dim=192 \end{tabular} &
\begin{tabular}{c} Patch Projection \\ R=2, Dim=192 \end{tabular} \\
\cline{3-5}
& & $\begin{bmatrix}\text{IPSA, head 4,}\\\text{CPSA, head 1,}\\\text{IPSA, head 4,}\\\text{patch. sz. 7 $\times$7}\end{bmatrix}$ $\times$ 1   & 
$\begin{bmatrix}\text{IPSA, head 6,}\\\text{CPSA, head 1,}\\\text{IPSA, head 6,}\\\text{patch. sz. 7 $\times$7}\end{bmatrix}$ $\times$ 1    &
$\begin{bmatrix}\text{IPSA, head 6,}\\\text{CPSA, head 1,}\\\text{IPSA, head 6,}\\\text{patch. sz. 7 $\times$7}\end{bmatrix}$ $\times$ 1    \\
\hline
\multirow{3}{*}{stage 3}  & \multirow{3}{*}{\begin{tabular}[c]{@{}c@{}}16$\times$\\\end{tabular}}  & 
\begin{tabular}{c} Patch Projection \\ R=2, Dim=256 \end{tabular} &
\begin{tabular}{c} Patch Projection \\ R=2, Dim=384 \end{tabular} &
\begin{tabular}{c} Patch Projection \\ R=2, Dim=384 \end{tabular} \\
\cline{3-5}
& & $\begin{bmatrix}\text{IPSA, head 8,}\\\text{CPSA, head 1,}\\\text{IPSA, head 8,}\\\text{patch. sz. 7 $\times$7}\end{bmatrix}$ $\times$ 3   & 
$\begin{bmatrix}\text{IPSA, head 12,}\\\text{CPSA, head 1,}\\\text{IPSA, head 12,}\\\text{patch. sz. 7 $\times$7}\end{bmatrix}$ $\times$ 3    &
$\begin{bmatrix}\text{IPSA, head 12,}\\\text{CPSA, head 1,}\\\text{IPSA, head 12,}\\\text{patch. sz. 7 $\times$7}\end{bmatrix}$ $\times$ 6    \\
\hline
\multirow{3}{*}{stage 4} & \multirow{3}{*}{\begin{tabular}[c]{@{}c@{}}32$\times$\\\end{tabular}}  &
\begin{tabular}{c} Patch Projection \\ R=2, Dim=512 \end{tabular} &
\begin{tabular}{c} Patch Projection \\ R=2, Dim=768 \end{tabular} &
\begin{tabular}{c} Patch Projection \\ R=2, Dim=768 \end{tabular} \\
\cline{3-5}
& & $\begin{bmatrix}\text{IPSA, head 16,}\\\text{CPSA, head 1,}\\\text{IPSA, head 16,}\\\text{patch. sz. 7 $\times$7}\end{bmatrix}$ $\times$ 1   & 
$\begin{bmatrix}\text{IPSA, head 24,}\\\text{CPSA, head 1,}\\\text{IPSA, head 24,}\\\text{patch. sz. 7 $\times$7}\end{bmatrix}$ $\times$ 1    &
$\begin{bmatrix}\text{IPSA, head 24,}\\\text{CPSA, head 1,}\\\text{IPSA, head 24,}\\\text{patch. sz. 7 $\times$7}\end{bmatrix}$ $\times$ 1    \\
\thickhline
\end{tabular}
\end{table*}

\section{Experiment}
\label{experiment}
We conduct image classification, object detection, and semantic segmentation experiments on ImageNet-1K\cite{deng2009imagenet}, COCO 2017\cite{lin2014microsoft}, and ADE20K\cite{zhou2017scene} respectively. In the following, we compare the three tasks between CAT architecture and state-of-the-arts architectures, then we report ablation experiments of some designs we adopted in CAT.
\subsection{Image Classification}
\paragraph{Details}For image classification, we report the top-1 accuracy with a single crop on ImageNet-1K\cite{deng2009imagenet}, which contains 1.28M training images and 50K validation images from 1000 categories. The setting in our experiments is mostly following \cite{touvron2020training}. We employ the batch size of 1024, the initial learning rate of 0.001, and the weight decay of 0.05. We train the model for 300 epochs with AdamW\cite{loshchilov2018fixing} optimizer, cosine decay learning rate scheduler, and linear warm-up of 20 epochs. stochastic depth\cite{huang2016deep} is used in our training, rate of 0.1, 0.2, and 0.3 for three variants architecture respectively, and dropout\cite{srivastava2014dropout} is adopted in self-attention of CAB with the rate of 0.2 to avoid overfitting. We use most of the regularization strategies and augmentation in \cite{touvron2020training} that similar to \cite{liu2021swin} to make our results more comparable and convincing.

\begin{table*}[h]
\small
\centering
\caption{The comparison of CAT with other networks on ImageNet. $\ddagger$ indecates that Swin-T without shifted window.}
\label{imamgenet}
\begin{tabular}{c|ccc|c}
\thickhline
 Model  & Resolution & Params(M) & FLOPs(B) & Top-1(\%) \\
\thickhline
\multicolumn{5}{c}{\textbf{CNN-based networks}}\\ 
\hline
ResNet50\cite{he2016deep} &$224\times224$ & 26 & 4.1 & 76.6\\
ResNet101\cite{he2016deep} &$224\times224$ & 45 & 7.9 & 78.2\\
\hline
X50-32x4d\cite{xie2017aggregated} & $224\times224$ & 25 & 4.3 & 77.9 \\
x101-32x4d\cite{xie2017aggregated} & $224\times224$ & 44 & 8.0 & 78.7 \\
\hline
EffifientNet-B4\cite{tan2019efficientnet} &$380\times380$ &19 &4.2 &82.9 \\
EffifientNet-B5\cite{tan2019efficientnet} &$528\times528$ &30 &9.9 &83.6 \\
EffifientNet-B6\cite{tan2019efficientnet} &$600\times600$ &43 &19.0 &84.0 \\
\hline
RegNetY-4G\cite{radosavovic2020designing} &$224\times224$ &21 &4.0 &80.0\\
RegNetY-8G\cite{radosavovic2020designing} &$224\times224$ &39 &8.0 &81.7\\
RegNetY-16G\cite{radosavovic2020designing} &$224\times224$ &84 &16.0 &82.9\\
\thickhline
\multicolumn{5}{c}{\textbf{Transformer-based networks}}\\ 
\hline
ViT-B/16\cite{dosovitskiy2020image} &$384\times384$ &86 &55.4 &77.9 \\
ViT-L/16\cite{dosovitskiy2020image} &$384\times384$ &307 &190.7 &76.5 \\
\hline
TNT-S\cite{han2021transformer} &$224\times224$ &24 &5.2 &81.3 \\
TNT-B\cite{han2021transformer} &$224\times224$ &66 &14.1 &82.8 \\
\hline
CrossViT-15\cite{chen2021crossvit} &$224\times224$ &27 &5.8 &81.5 \\
CrossViT-18\cite{chen2021crossvit} &$224\times224$ &44 &9.0 &82.5 \\
\hline
PVT-S\cite{wang2021pyramid} & $224\times224$ & 24.5 & 3.8 & 79.8 \\
PVT-M\cite{wang2021pyramid} & $224\times224$ & 44.2 & 6.7 & 81.2 \\
PVT-L\cite{wang2021pyramid} & $224\times224$ & 61.4 & 9.8 & 81.7 \\
\hline
$\text{Swin-T}^{\ddagger}$\cite{liu2021swin} &$224\times224$ &29 & - &80.2 \\
Swin-T\cite{liu2021swin} &$224\times224$ &29 &4.5 &81.3 \\
Swin-B\cite{liu2021swin} &$224\times224$ &88 &15.4 &83.3 \\
\thickhline
CAT-T(\textbf{ours}) &$224\times224$ & 17 & 2.8  & 80.3\\
CAT-S(\textbf{ours}) &$224\times224$ & 37 & 5.9  & 81.8\\
CAT-B(\textbf{ours}) &$224\times224$ & 52 & 8.9  & 82.8\\
\thickhline
\end{tabular}
\end{table*}

\paragraph{Results}In Table~\ref{imamgenet}, we present our experimental results, which demonstrates that our CAT-T could achieve the precision of 80.3\% top-1 when FLOPs were 65\% less than ResNet101\cite{he2016deep}. Meanwhile, the top-1 of our CAT-S and CAT-B on images with the resolution of $224\times224$ were 81.8\% and 82.8\%, respectively. Such a result is comparable with the results of state-of-the-arts in the Table For instance, compared with Swin-T\cite{liu2021swin}, which has a similar computation, our CAT-S has improved by 0.5\%. In particular, our method has a much stronger ability to catch the relationship between patches than shifted operation in Swin\cite{liu2021swin}. Swin-T(w. shifted) improves 1.1\% top-1 accuracy, and CAT-S surpasses 1.6\%.

\subsection{Object detection}

\paragraph{Details}For object detection, we conduct experiments on COCO 2017\cite{lin2014microsoft} with metric of mAP, which consists of 118k training, 5k validation, and 20k test images from 80 categories. We experiment on the some frameworks to evaluate our architecture. The batch size of 16, the initial learning rate of 1e-4, weight decay of 0.05 are used in our experiments. AdamW\cite{loshchilov2018fixing} optimizer, 1x schedule, and NMS\cite{neubeck2006efficient} are employed. Other settings are the same as MMDetection\cite{mmdetection}. Note that a stochastic depth\cite{huang2016deep} rate of 0.2 to avoid overfitting. About multi-scale strategy, we trained with randomly select one scale shorter side from 480 to 800 spaced by 32 while the longer side is less than 1333 same as \cite{carion2020end, sun2020sparse}. 

\begin{table*}[h]
\small
\centering
\caption{The comparison of CAT with other backbones with various methods on COCO detection. $\dagger$ indicates that trained with multi-scale strategy. FLOPs is evalutated on $800\times 1280$.}
\label{det}
\addtolength{\tabcolsep}{-4pt}
\resizebox{\textwidth}{!}{
\begin{tabular}{cc|ccc|ccc|cc}
\thickhline
 Method & Backbone & $AP^{box}$ & $AP^{box}_{50}$ & $AP^{box}_{75}$ & $AP^{mask}$ & $AP^{mask}_{50}$ & $AP^{mask}_{75}$ & Params(M) & FLOPs(G) \\
\thickhline
\multirow{4}{*}{Mask R-CNN\cite{he2017mask}} & 
ResNet50\cite{he2016deep} & 38.0 & 58.6 & 41.4 & 34.4 & 55.1 & 36.7 & 44 & 260\\ &
ResNet101\cite{he2016deep} & 40.4 & 61.1 & 44.2 & 36.4 & 57.7 & 38.8 & 63 & 336\\ &
$\text{CAS-S(\textbf{ours})}^\dagger$ & 41.6 & 65.1 & \textbf{45.4} & 38.6 & 62.2 & 41.0 & 57 & 295\\ &
$\text{CAS-B(\textbf{ours})}^\dagger$ & \textbf{41.8} & \textbf{65.4} & 45.2 & \textbf{38.7} & \textbf{62.3} & \textbf{41.4} & 71 & 356\\
\thickhline
 Method & Backbone & $AP^{box}$ & $AP^{box}_{50}$ & $AP^{box}_{75}$ & $AP^{box}_S$ & $AP^{box}_M$ & $AP^{box}_L$ & Params(M) & FLOPs(G) \\
\thickhline
\multirow{4}{*}{FCOS\cite{tian2019fcos}} & 
ResNet50\cite{he2016deep} & 36.6 & 56.0 & 38.8 & 21.0 & 40.6 & 47.0 & 32 & 201\\ &
ResNet101\cite{he2016deep} & 39.1 & 58.3 & 42.1 & 22.7 & 43.3 & 50.3 & 51 & 277\\ &
CAT-S\textbf{(ours)} & 40.0 & 60.7 & 42.6 & 24.5 & 42.7 & 52.4 & 45 & 245\\ &
CAT-B\textbf{(ours)} & \textbf{41.0} & \textbf{62.0} & \textbf{43.2} & \textbf{25.7} & \textbf{43.5} & \textbf{53.8} & 59 & 303\\
\hline
\multirow{4}{*}{ATSS\cite{zhang2020bridging}} & 
ResNet50\cite{he2016deep} & 39.4 & 57.6 & 42.8 & 23.6 & 42.9 & 50.3 & 32 & 205\\ &
ResNet101\cite{he2016deep} & 41.5 & 59.9 & 45.2 & 24.2 & 45.9 & 53.3 & 51 & 281\\ &
CAT-S\textbf{(ours)} & 42.0 & 61.6 & 45.3 & 26.4 & 44.6 & 54.9 & 45 & 243\\ &
CAT-B\textbf{(ours)} & \textbf{42.5} & \textbf{62.4} & \textbf{45.8} & \textbf{27.8} & \textbf{45.2} & \textbf{56.0} & 59 & 303\\
\hline
\multirow{4}{*}{RetinaNet\cite{lin2017focal}} & 
ResNet50\cite{he2016deep} & 36.3 & 55.3 & 38.6 & 19.3 & 40.0 & 48.8 & 38 & 234\\ &
ResNet101\cite{he2016deep} & 38.5 & 57.8 & 41.2 & 21.4 & 42.6 & 51.1 & 57 & 315\\ &
CAT-S\textbf{(ours)} & 40.1 & 61.0 & 42.6 & 24.9 & 43.6 & 52.8 & 47 & 276\\ &
CAT-B\textbf{(ours)} & \textbf{41.4} & \textbf{62.9} & \textbf{43.8} & \textbf{24.9} & \textbf{44.6} & \textbf{55.2} & 62 & 337\\
\hline
\multirow{5}{*}{Cascade R-CNN\cite{cai2018cascade}} & 
ResNet50\cite{he2016deep} & 40.4 & 58.9 & 44.1 & 22.8 & 43.7 & 54.0 & 69 & 245 \\ &
ResNet101\cite{he2016deep} & 42.3 & 60.8 & 46.1 & 23.8 & 46.2 & 56.4 & 88 & 311 \\ &
CAT-S\textbf{(ours)} & 44.1 & 64.3 & 47.9 & 28.2 & 46.9 & 58.2 & 82 & 270\\ &
CAT-B\textbf{(ours)} & 44.8 & 64.9 & 48.8 & 27.7 & 47.4 & 59.7 & 96 & 330\\ &
$\text{CAS-S(\textbf{ours})}^\dagger$ & 45.2 & 65.6 & 49.2 & 30.2 & 48.6 & 58.2 & 82 & 270\\ &
$\text{CAS-B(\textbf{ours})}^\dagger$ & \textbf{46.3} & \textbf{66.8} & \textbf{49.9} & \textbf{30.8} & \textbf{49.5} & \textbf{59.7} & 96 & 330 \\
\thickhline
\end{tabular}
}
\end{table*}

\paragraph{Results} As demonstrated in Table~\ref{det}, we used CAT-S and CAT-B as backbone in some anchor-based and anchor-free frameworks, both have better performance and comparable or fewer computational cost. CAT-S improves FCOS\cite{tian2019fcos} by 3.4\%, RetinaNet\cite{lin2017focal} by 3.7\%, and Cascade R-CNN\cite{cai2018cascade} by 4.8\% with multi-scale strategy. While for instance segmentation, we use the framework of MASK R-CNN\cite{he2017mask}, and the mask mAP improves 4.2\% with CAT-S. All methods we experimented on have better performance than the original, demonstrating our CAT has a better ability to be feature extraction.

\subsection{Semantic Segmentation}

\paragraph{Details}For semantic segmentation, we experiment on ADE20K\cite{zhou2017scene} which has 20k images for training, 2k images for validation, and 3k images for testing.  The setting is as follows, the initial learning rate is 6e-5, the batch size is 16 for 160k and 80k iterations in total, the weight decay is 0.05, and the warm-up iteration is 1500. We conduct experiments at the framework of Semantic FPN\cite{kirillov2019panoptic} with the input of $512\times512$, and using the basic setting in MMSegmentation\cite{mmseg2020}. Note that the stochastic depth\cite{huang2016deep} rate of 0.2 is used in CAT while training.

\begin{table*}[h]
\addtolength{\tabcolsep}{-3.0pt}
\small
\centering
\caption{Semantic segmentation performance on ADE20K. $\dagger$ indicates that the model is pre-trained on ImageNet-22k. $\ddagger$ indicates that trained with 80k iterations. FLOPs is evaluated on $1024 \times 1024$.}
\label{ade20k}
\begin{tabular}{cc|cc|c}
\thickhline
 Method & Backbone & Params(M) & FLOPs(G) & mIoU \\
\thickhline
DANet\cite{fu2019dual} & ResNet101\cite{he2016deep}  & 69 & 1119 & 45.0 \\
OCRNet\cite{yuan2019object} & ResNet101\cite{he2016deep} & 56 & 923 & 44.1\\
OCRNet\cite{yuan2019object} & HRNet-w48\cite{sun2019deep} & 71 & 664 & 45.7\\
DeeplabV3+\cite{chen2018encoder} & ResNet101\cite{he2016deep} & 63 & 1021 & 44.1\\
DeeplabV3+\cite{chen2018encoder} & ResNeSt-101\cite{zhang2020resnest} & 66 & 1051 & 46.9\\
UperNet\cite{xiao2018unified} & ResNet101\cite{he2016deep} & 86 & 1029 & 44.9\\
SETR\cite{xiao2018unified} & $\text{T-Large}^\dagger$ & 308 & - & 50.3\\
\hline
\multirow{4}{*}{Semantic FPN\cite{kirillov2019panoptic}} & ResNet50\cite{he2016deep} & 29 & 183 & 39.1 \\
                                                         & ResNet101\cite{he2016deep} & 48 & 260 & 40.7 \\
                                                         & CAT-S\textbf{(ours)} & 41 & 214 & 42.8\\
                                                         & CAT-B\textbf{(ours)} & 55 & 276 & 44.9\\
\hline
\multirow{4}{*}{$\text{Semantic FPN}^\ddagger$} & ResNet50\cite{he2016deep} & 29 & 183 & 36.7 \\
                                                & ResNet101\cite{he2016deep} & 48 & 260 & 38.8 \\
                                                & CAT-S\textbf{(ours)} & 41 & 214 & 42.1 \\
                                                & CAT-B\textbf{(ours)} & 55 & 276 & 43.6 \\
\thickhline
\end{tabular}
\end{table*}

\paragraph{Results} As shown in Table~\ref{ade20k}, we employ CAT-S and CAT-B as the backbone, with the framework of Semantic FPN\cite{xiao2018unified}. Semantic FPN achieves better performance with CAT-S and CAT-B, especially, we achieve 44.9\% mIoU with 160k iterations and CAT-B, 4.2\% improved compared ResNet101\cite{he2016deep} as the backbone, making Semantic FPN obtains comparable performance as other methods, while for 80k iterations, result is enhanced 4.8\%, which illustrates that our architecture is more powerful than ResNet\cite{he2016deep} to be a backbone.

\subsection{Ablation Study}
\label{ablation-study}
In this section, we report results of the ablation experiments for some designs we made in designing the architecture and in conducting the experiments on ImageNet-1K\cite{deng2009imagenet}, COCO 2017\cite{lin2014microsoft}, and ADE20K\cite{zhou2017scene}.
\paragraph{Patch Embedding function}We compare the embedding function in patch embedding layer, convolutional method and method in \cite{yolov5}, the former conduct convolutional layer with the kernel size of $4\times4$ and stride of 4 to reduce the resolution of input to $1/4$ of origin, the latter slice the input from $H\times W \times H \times C$ to $H/S \times W/S \times SC$, where S in ours is 4 to implement the same as the former. The results in Table~\ref{ablation} show that the two methods have same performance. To better compare with other work\cite{liu2021swin}, we choose the convolutional method as the default setting.

\paragraph{Multi-head and shifted window} Multi-head is proposed in \cite{vaswani2017attention}, which represents different semantic features among words. We set the number of heads equal to patch size in each CPSA, which is useless to the performance, presented in Table~\ref{ablation}. To study the shifted window in Swin\cite{liu2021swin}, we also experimented  w./wo. the shifted window at the third block of CAB, the result shows that the shifted operation does not perform better in our architecture.

\begin{minipage}{\textwidth}
   \begin{minipage}[t]{0.4\textwidth}
     \small
     \centering
     \makeatletter\def\@captype{table}\makeatother\caption{Ablation study on multi-head in CPSA, shifted window in second IPSA block in CAB, and slice or convlutional method in patch embedding layer, using CAT-S architecture on ImageNet-1K.}
     \label{ablation}
     \addtolength{\tabcolsep}{-3.5pt}
        \begin{tabular}{cccc|c} 
              \thickhline
              \multicolumn{1}{c}{multi-head}  &\multicolumn{1}{c}{shifted} & \multicolumn{1}{c}{slice} & \multicolumn{1}{c}{conv.}& Top-1(\%)\\
              \midrule
               & \checkmark &  & \checkmark & 81.7 \\
              \checkmark &  & & \checkmark & 81.6 \\
               &  & \checkmark & & \textbf{81.8} \\
               &  &  & \checkmark & \textbf{81.8} \\
              \thickhline
      \end{tabular}
  \end{minipage}
  \hspace{2mm}
  \begin{minipage}[t]{0.58\textwidth}
    \small
    \centering
    \makeatletter\def\@captype{table}\makeatother\caption{Ablation study on the absolute position encoding and dropout in self-attention of CPSA on three benchmarks with CAT-S architecture. FCOS\cite{tian2019fcos} with 1x schedule on COCO 2017 and Semanticc FPN\cite{xiao2018unified} with 80k iterations on ADE20K is used. attn.d: dropout of selt-attention. abs.pos.: absolute position encoding.}
    \label{ablation2}
    \addtolength{\tabcolsep}{-3.5pt}
      \begin{tabular}{c|cc|ccc|c}
        \thickhline
         & \multicolumn{2}{c|}{ImageNet} & \multicolumn{3}{c|}{COCO 2017} & \multicolumn{1}{c}{ADE20k} \\
         & top-1 & top-5  & AP & AP$^\text{50}$ & AP$^\text{75}$ & mIoU \\
        \hline
        no attn.d & 81.5 & 95.2 & 39.8 & 60.5 & 43.0 & 42.0 \\
        attn.d 0.2 & \textbf{81.8} & \textbf{95.6} & \textbf{40.0} & \textbf{60.7} & \textbf{43.2} & \textbf{42.1} \\
        \hline
        no abs.pos. & 81.6 & 95.3 & 39.6 & 60.2 & 42.9  & 41.8 \\
        abs.pos. & \textbf{81.8} & \textbf{95.6} & \textbf{40.0} & \textbf{60.7} & \textbf{43.2} & \textbf{42.1} \\
        \thickhline
      \end{tabular}
  \end{minipage}
\end{minipage}

\paragraph{Absolute position and dropout in self-attention of CPSA} We conduct ablation study on absolute position encoding for CPSA, and it improves the performance on three benchmarks. To better training, we adopted the dropout\cite{srivastava2014dropout} of self-attention in CPSA and set the rate of 0.0 and 0.2. The rate of 0.2 achieves the best performance, illustrating there is a little overfitting in CPSA. All results are reported in Table~\ref{ablation2}.

\section{Conclusion}
\label{conclution}
In this paper, the proposed Cross Attention is proposed to better combine the virtue of local feature extraction in CNN with the virtue of global information extraction in Transformer, and build a robust backbone, which is CAT. It can generate features at different scales similar to most CNN-based networks, and it can also adapt to different sizes of inputs for other vision tasks. CAT achieves state-of-the-arts performance on various vision task datasets (e.g., ImageNet-1K\cite{deng2009imagenet}, COCO 2017\cite{lin2014microsoft}, ADE20K\cite{zhou2017scene}). The key is that we alternate attention inner the feature map patch and attention on the single-channel feature map without quite increasing the computation to capture local and global information. We hope that our work will be a step in the direction of integrating CNN and Transformer to create a multi-domain approach.

\bibliographystyle{unsrt}
\bibliography{cat}

\end{document}